\documentclass[11pt]{article}

\usepackage[letterpaper, textwidth=5.5in, textheight=9in, centering]{geometry}

\usepackage{mathptmx}
\usepackage[T1]{fontenc}
\usepackage[utf8]{inputenc}

\usepackage{microtype}

\usepackage{amsmath}
\usepackage{bm}

\usepackage{booktabs}
\usepackage{graphicx}
\graphicspath{{./}}

\usepackage{natbib}

\usepackage{hyperref}

\usepackage{xcolor}

\newcommand{\mat}[1]{\bm{#1}}
\newcommand{\vct}[1]{\bm{#1}}
\newcommand{\set}[1]{\mathcal{#1}}
\newcommand{\Reg}{\mathsf{Reg}}
\newcommand{\rglobal}{\pi_{\mathsf{global}}}
\newcommand{\rskill}{\pi_{\mathsf{skill}}}
\newcommand{\roracle}{\pi_{\mathsf{oracle}}}
\newcommand{\dsk}{\Delta_{\mathsf{sk}}}
\newcommand{\dor}{\Delta_{\mathsf{or}}}
\newcommand{\Ones}{\bm{1}}

\renewenvironment{abstract}{%
  \vspace{1ex}%
  \centerline{\large\bfseries Abstract}%
  \vspace{0.5ex}%
  \begin{quote}\small
}{%
  \end{quote}%
}

\newcommand{\keywords}[1]{%
  \par\noindent\small\textbf{Keywords:} \textit{#1}%
}

\title{\textbf{When Should Agent Trust Be Conditional?}\\[0.4ex]
  \large Characterizing and Attacking Skill-Conditional Reputation in Agent Swarms}

\author{%
Yihan Xia\\
Shenzhen University, China\\
\texttt{xiayihan2023@email.szu.edu.cn}
\and
Taotao Wang\thanks{Corresponding author.}\\
Shenzhen University, China\\
\texttt{ttwang@szu.edu.cn}
}
\date{}

\begin{document}

\maketitle

%% ---------------------------------------------------------------------------
\begin{abstract}
Open platforms increasingly route tasks among heterogeneous LLM agents---differing
in base model, scaffold, and tool stack---whose competence varies sharply by skill:
an agent excellent at one skill may be useless at another. The standard reputation
approach summarizes each agent by a single global trust score, but that scalar is
the wrong object here, because routing every task to the globally most-trusted agent
leaves the value of specialization unclaimed. We study \emph{skill-conditional trust}
$R(i\mid k)$---the trust to place in agent $i$ for a task requiring skill $k$,
rather than one score per agent---and pose three falsifiable questions: \emph{when} is conditioning worth it, \emph{how much} cross-skill evidence
should be borrowed, and \emph{whether} that borrowing is safe.
A controlled phase-diagram analysis answers the first two: conditional trust
wins only in a specific regime---high agent heterogeneity, sparse per-skill
evidence, and correlated skills---and the coupling strength $\beta$ that buys
this data efficiency is \emph{dual-use}, because the same cross-skill borrowing
is also a laundering channel. On a public benchmark of $14$ genuinely
heterogeneous AppWorld agents, real pools land inside the beneficial
regime---a small but genuine gain, with the per-skill best agent genuinely
changing across skills. We then show that an attacker with cheap evidence in one
skill and none in a target skill hijacks the conditional router, driving
routing regret from $0$ to $0.94$ on a pool our zero-cost Conditional Information
Value Test (CIVT) rates \textsc{green}---while the ungated trust verdict it contaminates
reads $-0.06$ instead of the honest $+0.19$. A zero-evidence gate bounds the
attack but does not eliminate it; we characterize the residual cost under an
explicit budget. We do \emph{not} claim Sybil-resistance---we quantify the
trade-off.
\end{abstract}

\keywords{multi-agent systems, LLM agents, trust and reputation, task routing,
adversarial robustness, reputation laundering, estimation under sparsity}

%% ===========================================================================
\section{Introduction}
\label{sec:intro}

Open platforms are increasingly populated by \emph{heterogeneous LLM agents}
offered as services---each a distinct combination of base model, scaffold, and
tool stack---and standards for inter-agent delegation and trust are already being
drafted around them~\cite{interagenttrust2025,trustworthyagents2025survey}. An
orchestrator must decide which agent to trust with each incoming task. The
dominant approach to such trust, inherited from reputation systems for online
marketplaces and peer-to-peer networks~\cite{kamvar2003eigentrust}, summarizes
each participant by a \emph{single global score}: how much the system, on
aggregate, should trust them. That scalar is exactly right when a participant's
trustworthiness is one-dimensional.

But these agents are not uniformly good, and \emph{which} one is best depends on
the task---a regularity that model- and agent-routing systems already exploit to
trade cost against quality~\cite{ong2024routellm,li2026routingsurvey}. On the
public AppWorld benchmark~\cite{trivedi2024appworld} we study
(\S\ref{sec:realdata}), the agent that is best at \emph{phone} tasks (an
\textsc{IPFunCall} scaffold on GPT-4 Turbo) is not the agent best at \emph{venmo}
payments (a \textsc{PlanExec} scaffold on GPT-4o), which is in turn not the best
at \emph{spotify} (FullCodeRefl on GPT-4o). A single global trust score
implicitly assumes an agent is uniformly trustworthy across tasks; for
skill-specialized agents that assumption is simply false, and routing every task
to the globally most-trusted agent forfeits the value of specialization.

We therefore ask whether the trust \emph{object} itself should be
\emph{conditional}: not one score per agent, but one score per agent
\emph{per skill}---$R(i\mid k)$, the trust to place in agent $i$ when the task
requires skill $k$. The difficulty is that per-skill evidence is
sparse---an agent may have run only a handful of episodes on any given skill---so
a naive per-skill estimate is too noisy to route on. The natural remedy is to
\emph{borrow} evidence across correlated skills: an agent's record on related
skills is informative about a skill it has rarely attempted---an agent reliable at
\emph{venmo} payments is a reasonable prior for the closely related
\emph{splitwise}, even with few direct splitwise episodes. This borrowing of
strength across related cells is the classic variance-reduction logic of
empirical-Bayes shrinkage~\cite{james1961estimation,efron1977stein}, and the same
move classical reputation systems use to cope with sparse direct
evidence---structured borrowing along a correlation graph, applied here on the
agent--skill axis rather than the peer graph (\S\ref{sec:rel-eigentrust}).

\paragraph{The question, sharpened.}
It is tempting to stop there and assert that ``conditional trust must be
better.'' It is not. Earlier attempts---including our own---to claim blanket
superiority of conditional routing do not survive scrutiny: when agents are
\emph{not} specialized, or when per-skill evidence is plentiful, a single
global-best agent is genuinely hard to beat, and cross-skill borrowing can even
inject harmful bias. We therefore reframe the contribution from \emph{advocacy}
to \emph{characterization}, and ask three falsifiable questions:
\textbf{when} does conditioning on a task's skill beat a single global-best
agent; \textbf{how much} cross-skill evidence should one borrow when per-skill
data is sparse; and \textbf{whether} that borrowing opens an attack surface.
The third question is not an afterthought: the very mechanism that buys data
efficiency---cross-skill evidence borrowing, tuned by a coupling strength
$\beta$---turns out to be a reputation-laundering channel, echoing how added
inter-agent trust can enlarge the attack surface rather than only improve
collaboration~\cite{xu2025trust,prakash2026provenance}. \emph{How much} to borrow
and \emph{whether} it is safe are thus two faces of one knob. This
measurement-first, mechanism-then-attack stance is the honest and, we argue, the
more durable contribution.

\paragraph{What we find.}
The three answers turn out sharper than the framing promises. \emph{When}:
conditioning helps only in a specific regime---high heterogeneity, sparse
per-skill evidence, and correlated skills---and a zero-cost test (CIVT) reads off
from existing logs, before any model is run, whether a given pool lies inside it;
real public pools land just inside, with a small but genuine gain. \emph{How
much}: the borrowing strength has a constrained optimum, since pushing it past
$\beta\!\approx\!0.1$ adds little accuracy while widening the laundering channel.
\emph{Whether it is safe}: here the result is alarming. On a pool CIVT
\emph{certifies} as carrying real conditional value, an attacker with perfect
evidence on one cheap skill and \emph{no} evidence on the target skill captures
the router for the price of a single fabricated episode, driving routing regret
from $0$ to $0.94$ (\S\ref{sec:attack}); the same contamination flips the trust
verdict a defender reads from an honest $+0.19$ (``conditioning is clearly worth
it'') to $-0.06$ (``conditioning is harmful''). The defense that should work makes
it worse: the adaptive coupling that gracefully avoids over-borrowing on benign
pools rides the very correlation the attacker exploits, so robustness at low skill
correlation \emph{inverts} into vulnerability at high correlation. Rate limits and
coupling caps are algebraically powerless too; only a structural rule that refuses
to borrow for an agent with no direct target evidence bounds the attack, and even
that is bypassable at a cost we quantify (\S\ref{sec:defense}).

\paragraph{Contributions.}
We answer the three questions in four steps---formalizing the object, mapping
when and how much to condition, placing real data on that map, and then turning
the borrowing mechanism against itself as an attack:
\begin{enumerate}
\item \textbf{A skill-conditional trust object} $R(i\mid k)$, with idealized
routers ($\rglobal$, $\rskill$, $\roracle$) and a zero-cost de-risking test
(CIVT) that decides, from existing logs alone, whether skill conditioning carries
value (\S\ref{sec:formulation}).
\item \textbf{A phase diagram of when/how-much/whether} (\S\ref{sec:mechanism}):
conditioning wins only under high heterogeneity $H$ + sparse evidence $N$ +
correlated skills $C$; the borrowing strength $\beta$ has a \emph{constrained}
optimum because the borrowing channel is dual-use; and the coupling must be
\emph{adaptively estimated}, not assumed.
\item \textbf{Real-data placement} (\S\ref{sec:realdata}): on 14 heterogeneous
public AppWorld agents, real pools land in the beneficial regime (validated on
both continuous and binary outcomes); across five landing points the conditional
gain tracks skill correlation $C$ as the phase diagram predicts, and CIVT
\emph{discriminates}---\textsc{green} on \texttt{test\_normal}, \textsc{amber} on
the harder \texttt{test\_challenge}---rather than rubber-stamping.
\item \textbf{Adversarial analysis} (\S\ref{sec:attack}): cross-skill laundering
hijacks the conditional router at small $\beta$; we give a budgeted threat model
and defenses that bound---not eliminate---attack benefit.
\end{enumerate}

%% ===========================================================================
\section{Related Work}
\label{sec:related}

Our work sits at the intersection of four threads: the reputation paradigm we
depart from (and its classical Sybil/whitewashing attacks, which we specialize);
LLM-agent routing, to which we add an adversarial trust dimension; the emerging
agent-trust-and-security literature, which we make \emph{quantitative}; and
classical shrinkage estimation, which we re-read as a security knob. We position
against each in turn.

\paragraph{Spectral reputation aggregation and its classical attacks.}
Power-iterating a normalized adjacency matrix to its principal eigenvector is
the canonical recipe for ranking participants in an open graph:
\textsc{PageRank} applies it to web authority~\cite{page1999pagerank},
\textsc{EigenTrust} to peer-to-peer reputation~\cite{kamvar2003eigentrust}, and
a long line of work since studies the recipe's robustness, its pretrust/damping
defenses, and personalized or context-aware variants. What this family shares is
the \emph{object} it produces: a single scalar score per participant (in
personalized variants, per \emph{evaluator}). The classical threat models are the Sybil
attack~\cite{douceur2002sybil}---forging many identities to outvote honest
participants---and the cheap-pseudonym or \emph{whitewashing}
problem~\cite{friedman2001whitewashing}, shedding a bad history by re-registering.
Both are provably hard to eliminate in open systems: \citet{cheng2005sybilproof}
prove that \emph{no} symmetric reputation function---a class that includes
\textsc{PageRank}-style scores---can be Sybil-proof, so every such mechanism
admits a beneficial Sybil attack. This impossibility is exactly why
classical defenses are \emph{budgeted}---bounding rather than eliminating harm---as
in the social-network admission control of \textsc{SybilGuard} and
\textsc{SybilLimit}~\cite{yu2006sybilguard,yu2008sybillimit}.
We depart on the conditioning \emph{axis}, not the aggregation rule: we make the
trust object depend on the \emph{task's skill}---which agent to trust depends on
what is being asked---recovering the single global score as the no-conditioning
special case (\S\ref{sec:rel-eigentrust}). Our attacks (\S\ref{sec:attack}) are
correspondingly cross-skill reputation-laundering variants of Sybil/whitewashing,
and we inherit the classical impossibility---our defenses bound the attack benefit
under an explicit budget but are not Sybil-proof.

\paragraph{LLM-agent routing and model selection.} A large recent literature
routes queries among models or agents to cut cost or raise quality: learned
cost-routers such as \textsc{RouteLLM}~\cite{ong2024routellm}, cost-cutting model
cascades such as \textsc{FrugalGPT}~\cite{chen2023frugalgpt}, dynamic
routing/cascading systems surveyed in~\cite{li2026routingsurvey}, cost-aware
cross-attention routers~\cite{crossattn2025routing}, and mixture-of-experts
architectures~\cite{jiang2024mixtral} that route within a single model. This line
has largely matured from a research question into infrastructure, and it is not
our competitor: it routes on \emph{cost} or \emph{claimed capability}, assumes the
candidate agents are non-strategic, and presupposes a trustworthy quality signal.
We study the prior question that this presupposition hides---when, and how safely,
a \emph{trust} signal should be conditioned on skill at all. Without a reliable,
attack-resistant reputation signal, skill-routing is building on sand; we
characterize that signal's value and its adversarial limits rather than proposing
a better router.

\paragraph{Agent trust, reputation, and security.} As the ``agentic web'' begins
to deploy, trust between agents has become an active concern, and recent work
approaches our thesis closely. Industry-facing protocol studies catalog inter-agent
trust mechanisms across emerging standards (A2A, AP2, ERC-8004)
\cite{interagenttrust2025}, and surveys map the threat landscape of trustworthy
LLM agents~\cite{trustworthyagents2025survey} and zero-trust agentic
architectures~\cite{zerotrust2025agentic}; these establish that attack-resistant
agent reputation is a real, deployed open problem but remain framework-level. Two
papers are close enough to require sharp differentiation.
\citet{prakash2026provenance} identify a \emph{provenance paradox}: self-claimed
quality in multi-agent LLM delegation collapses routing below random, and their
remedy is protocol-level attestation of identity and delegation contracts. Our
launderer is strictly harder to defend against, because it does not lie: it holds
\emph{genuine} observed performance on a farm skill and exploits the
$\beta$-coupling of Eq.~\eqref{eq:pool} to contaminate the target estimate---the
signal is real and attestable, the \emph{pooling structure} is the vulnerability,
so attestation does not address it. \citet{xu2025trust} document a
\emph{trust paradox}---parameterized inter-agent trust improves collaboration while
enlarging the attack surface---and their trust parameter plays the role of our
$\beta$; but they treat trust as undifferentiated information exposure, whereas we
give a \emph{skill-conditional} structure and quantify the cost as routing regret,
a phase diagram of when conditioning pays, and a bounded-attack benefit.
Closest in object are concurrent decentralized-trust proposals.
\citet{qi2025blockchain} equip each agent with a per-skill \emph{capability vector}
beside a scalar reputation and match tasks by capability, and \citet{wang2025evtrust}
make a decentralized trust mechanism evolutionarily stable under adversarial
mutation. Both \emph{build a mechanism}; neither asks our question. A per-skill
capability vector is precisely the object we estimate---but they do not analyze the
cross-skill \emph{borrowing} that sparse per-skill evidence forces: its dual-use as
a laundering channel, the budget degeneracy that makes the attack one-episode cheap,
or the regime in which conditioning carries value at all. Our contribution is the
conditions under which estimating that vector \emph{conditionally} helps, how much
to borrow, and what the borrowing costs under attack. Across this thread the work
either identifies a paradox qualitatively or proposes a trust \emph{mechanism}; our
contribution is the quantitative map---\emph{when} conditioning has value,
\emph{how much} borrowing is optimal, and the \emph{upper bound} on what an attacker
gains---rather than a new framework.

\paragraph{Estimation under sparsity.} Borrowing strength across related cells to
reduce variance at the cost of bias is classical: James--Stein
estimation~\cite{james1961estimation} and empirical-Bayes
shrinkage~\cite{efron1977stein} formalize when pooling beats independent
per-cell estimates. Our coupling-weighted estimator (Eq.~\eqref{eq:pool}) is an
instance of this family on the agent--skill matrix, with $\beta$ controlling the
degree of pooling. The classical analysis treats $\beta$ as a purely statistical
bias--variance knob; our contribution is to give it a \emph{security} reading---the
same parameter that optimally trades bias for variance under honest data is the
parameter an attacker tunes to launder reputation, so its safe setting is bounded
from above by adversarial, not just statistical, considerations.

%% ===========================================================================
\section{Problem Formulation and Method}
\label{sec:formulation}

This section builds the machinery the rest of the paper measures and attacks. We
first explain why the live question is \emph{conditioning} rather than transitive
aggregation (\S\ref{sec:rel-eigentrust}). We then state the observation model, give a \emph{single}
coupling-weighted estimator whose four special cases are the methods we compare,
and define the routing regret that scores them. Finally we introduce CIVT---a
zero-cost test of whether conditioning carries value on a given log---and the
three pool statistics $(H,N,C)$ that the mechanism analysis of
\S\ref{sec:mechanism} sweeps.

\subsection{Why conditioning, not transitive aggregation}
\label{sec:rel-eigentrust}
Classical reputation systems confront \emph{sparse pairwise evidence}---a
participant has interacted with few others---and the canonical response is
\emph{transitive} borrowing across the graph: power-iterate a normalized trust
matrix to a fixed point, the same spectral recipe that ranks web pages
(\textsc{PageRank}~\cite{page1999pagerank}) and peers
(\textsc{EigenTrust}~\cite{kamvar2003eigentrust}). Our setting differs in two
ways that make \emph{conditioning}, not transitive aggregation, the live
question.

First, the sparse evidence lives on a different axis. Trust here is
\emph{competence}, and what is scarce is episodes in the agent$\times$\emph{skill}
matrix (an agent has few runs on a given skill). We therefore borrow evidence
\emph{across correlated skills}, along the structure encoded by the coupling
matrix $\mat{W}$ (Eq.~\eqref{eq:pool}), with the no-conditioning special case
(all-ones $\mat{W}$) recovering a single global score. This borrowing channel is
\emph{dual-use} in exactly the way transitive borrowing is in classical
systems---the strength $\beta$ that buys data efficiency is the same knob an
attacker tunes to launder reputation across skills (\S\ref{sec:attack}).

Second, outcomes here are \emph{program-verifiable} rather than peer-reported.
In that setting, a relay's identity matters when the object being relayed is an
assessment whose credibility must itself be judged. Here, the object is an
environment-verified episode record. Relays may still matter for discovery,
synchronization, or selective withholding of records, but once a record is
observed, its evidentiary weight does not depend on the peer who forwarded it.
The sparse axis therefore moves from who-trusts-whom evidence on a peer graph to
per-cell evidence in the agent$\times$\emph{skill} matrix, where a cell is one
agent--skill entry. Transitive reputation machinery becomes relevant again when
the signal includes unverifiable peer assessments, subjective ratings, or
cross-organization reports. In the
verifiable-outcome slice studied here, the estimation problem is how much
evidence to \emph{borrow across skills}.

\subsection{Setup and observation model}
\label{sec:setup}
A pool of $M$ agents serves tasks spanning $K$ skills. Agent $i$ has unknown
ground-truth competence $\theta_{i,k}\in[0,1]$ on skill $k$. At the
task level, an episode is an attempt by agent $i$ on task $t$, with skill label
$k(t)$ and verifiable outcome $y_{i,t}\in[0,1]$; for estimation we aggregate
these task-level records into agent--skill cells. Each cell $(i,k)$ is
observed through $n_{i,k}$ episodes with \emph{verifiable} outcomes
(program-checked pass/fail, or a graded score), summarized by an observed mean
score $o_{i,k}\in[0,1]$ carrying evidence mass $n_{i,k}$. Verifiability is the
feature that distinguishes our setting from classical reputation: an episode's
outcome is checked by the environment, not reported by a peer, so a single
observation $o_{i,k}$ is itself trustworthy and the only scarce resource is the
\emph{number} of observations $n_{i,k}$. This is why the live problem is
conditioning, not transitive aggregation (\S\ref{sec:rel-eigentrust}). Our object
of interest is the \emph{skill-conditional trust} $R(i\mid k)$: the degree to which
the orchestrator should trust agent $i$ on a task of skill $k$. Its ideal value is
the (unobserved) ground-truth competence $\theta_{i,k}$; the estimators of
\S\ref{sec:estimator} approximate it from the sparse, verifiable evidence
$\{o_{i,k},n_{i,k}\}$, and we write $\hat\tau_{i,k}$ for such an estimate.
Recovering $R(i\mid k)$ well---and, since routing selects the per-skill
$\arg\max$, recovering for each skill $k$ an across-agent \emph{ordering} close to
that of the true $\theta_{\cdot,k}$---is the core problem. (One may further
condition on role $r$ and query $q$; we study $k$ and leave $r,q$ to future work.)

\subsection{A single coupling-weighted estimator}
\label{sec:estimator}
All trust estimators we consider are instances of \emph{one} rule---evidence-mass
pooling across skills through a coupling matrix $\mat{W}$ ($K{\times}K$,
nonnegative, unit diagonal):
\begin{equation}
\hat\tau_{i,k}(\mat{W}) \;=\;
  \frac{\sum_{k'} W_{k,k'}\, n_{i,k'}\, o_{i,k'}}
       {\sum_{k'} W_{k,k'}\, n_{i,k'}} .
\label{eq:pool}
\end{equation}
Skill $k$ borrows skill $k'$'s evidence in proportion to the coupling $W_{k,k'}$
and to its evidence mass $n_{i,k'}$---an empirical-Bayes--style partial
pooling~\cite{james1961estimation,efron1977stein}. The methods differ \emph{only} in $\mat{W}$:
\begin{itemize}
\item \textbf{Independent} ($\mat{W}{=}\mat{I}$): $\hat\tau_{i,k}=o_{i,k}$. No
borrowing; unbiased but high-variance when $n_{i,k}$ is small.
\item \textbf{Global} ($\mat{W}{=}\Ones\Ones^{\top}$, all ones): pools every
skill into one score per agent, broadcast across $k$---the single global-score
object. Low variance, but biased when agents are skill-specialized.
\item \textbf{Conditional} (fixed block coupling): $W_{k,k'}{=}\beta$ if
$k\!\ne\!k'$ lie in the same correlated skill block, else $0$, with strength
$\beta\in[0,1]$. Independent is the limit $\beta{=}0$; the strength interpolates
between no pooling and within-block pooling.
\item \textbf{Adaptive}: data-estimated coupling
$W_{k,k'}{=}\beta\max(R_{k,k'},0)$, $k\!\ne\!k'$, with $\mat{R}$ the cross-agent
column-correlation matrix~\eqref{eq:corr}. Borrowing follows
\emph{empirically positive} correlation only; when skills are uncorrelated
$\mat{W}\!\to\!\mat{I}$ and it degrades gracefully to Independent (no backfire).
\end{itemize}
\begin{equation}
R_{k,k'}=\frac{\sum_i \tilde o_{i,k}\,\tilde o_{i,k'}}
              {\sqrt{\textstyle\sum_i \tilde o_{i,k}^2}\,
               \sqrt{\textstyle\sum_i \tilde o_{i,k'}^2}},
\qquad \tilde o_{i,k}=o_{i,k}-\bar o_{\cdot,k}.
\label{eq:corr}
\end{equation}
Here $\bar o_{\cdot,k}=\tfrac{1}{M}\sum_i o_{i,k}$ is the mean of skill column
$k$, so $R_{k,k'}$ is the across-agent Pearson correlation between skill columns
$k$ and $k'$. The coupling strength $\beta$ is thus a single knob trading variance
for bias---and, as \S\ref{sec:attack} shows, the same knob an attacker exploits.

\subsection{Conditional routing and regret}
Given estimates $\hat\tau$ and a candidate pool $\set{P}$, the conditional router
sends a skill-$k$ task to $a^\star_k=\arg\max_{i\in\set{P}}\hat\tau_{i,k}$. We
score a router by \emph{routing regret}, the normalized shortfall in \emph{true}
competence against the per-skill oracle:
\begin{equation}
\Reg(\hat\tau)=\frac{1}{K}\sum_{k=1}^{K}
  \Big(\max_{i\in\set{P}}\theta_{i,k}-\theta_{a^\star_k,\,k}\Big).
\label{eq:regret}
\end{equation}
Thus, estimates determine the routing decision, while regret is evaluated using
the true competence $\theta$.

\subsection{The Conditional Information Value Test (CIVT)}
\label{sec:civt}
Before deploying any estimator, we ask whether conditioning could
help \emph{at all} on a given evaluation log of verified episodes over agents and
skills. We answer with the \emph{Conditional Information Value Test}
(CIVT)---a zero-cost screen of whether the skill information is worth conditioning
on, computed from existing logs with no model run. Treating the observation matrix
$\mat{O}$ (entries $o_{i,k}$) as the
trust signal, we compare three idealized routers by their value $V(\cdot)$ (mean
score of the routed agent): $\rglobal$ (one fixed agent---the row of largest
overall mean, i.e.\ the single global-score baseline), $\rskill$ (the best agent
\emph{per skill}), and $\roracle$ (the best agent \emph{per task}, an upper
bound). The decision gaps are
\begin{align}
\dsk &= V(\rskill) - V(\rglobal)
  && \text{(value of conditioning)},\nonumber\\
\dor &= V(\roracle) - V(\rglobal)
  && \text{(total headroom).}
\label{eq:gaps}
\end{align}
A pool is labelled \textsc{green} (conditioning carries value) when
$\dor\!\ge\!0.05$, $\dsk\!\ge\!0.03$, and the per-skill best agent is not
unique. The numerical thresholds are practical-effect cutoffs on a
$[0,1]$ outcome scale, not fitted parameters or statistical significance tests:
$\dor\!\ge\!0.05$ requires at least five points of total routing headroom, while
$\dsk\!\ge\!0.03$ requires the deployable skill-level router to recover at least
three points. We report the raw gaps throughout, so changing these cutoffs changes
only the \textsc{green}/\textsc{amber} label, not the measured effect. Each
condition rules out a distinct failure of conditioning: $\dor$
demands that \emph{some} per-task headroom exists at all; $\dsk$ demands that a
\emph{skill-level} router (the deployable object, since per-task routing requires
an oracle) captures enough of it to be worth the machinery; and the
non-uniqueness condition rules out the degenerate case in which one agent happens
to win every skill, where a single global pick is already optimal. CIVT runs on
existing logs at zero model cost---it is the operational form of the \emph{when}
question (Proposition~1, \S\ref{sec:mechanism}), a de-risking step that decides
whether the estimators of \S\ref{sec:estimator} are even worth deploying before
any model is run.

\subsection{Characterizing a pool: $H$, $N$, $C$}
Whether conditioning pays off, and whether borrowing is safe, is governed by
three measurable properties of the observation matrix $\mat{O}$, whose row
$\vct{o}_i$ is agent $i$'s skill profile:
\begin{itemize}
\item \textbf{Heterogeneity} $H=\tfrac{1}{\sqrt{K}}\operatorname{mean}_{i<j}
  \lVert \vct{o}_i-\vct{o}_j\rVert_2$: how differently agents perform.
\item \textbf{Evidence} $N=\operatorname{mean}_{i,k} n_{i,k}$: episodes per cell
  (sparsity).
\item \textbf{Correlation} $C=\operatorname{mean}_{k\ne k'} R_{k,k'}$: how
  correlated the skill columns are (legitimacy of borrowing).
\end{itemize}
Sections~\ref{sec:mechanism}--\ref{sec:attack} show conditioning wins in a
specific $(H,N,C)$ regime, and that $C$ governs whether borrowing helps or
backfires.

%% ===========================================================================
\section{Mechanism: When, How Much, and Whether to Condition}
\label{sec:mechanism}

Real data answers \emph{where} a given pool lands but cannot vary $H$, $N$, and
$C$ independently to expose the underlying mechanism. We therefore isolate the
mechanism in a controlled data-generating process (DGP) in which all three are
knobs: $M{=}12$ agents and $K{=}4$ skills arranged in two correlated blocks, so
that the within-block correlation directly sets $C$, per-agent profiles set $H$,
and the episode count per cell sets $N$. We instantiate the estimator family of
Eq.~\eqref{eq:pool}---Global ($\mat{W}{=}\Ones\Ones^{\top}$), Independent
($\mat{W}{=}\mat{I}$), and Conditional (block coupling $\beta$)---sweep $H$, $N$,
$C$, and $\beta$ over $500$ trials per cell, and score each choice of $\mat{W}$ by
routing regret~\eqref{eq:regret} against the known ground truth. The three
propositions below answer, in order, the \emph{when}, \emph{how much}, and
\emph{whether} of \S\ref{sec:intro}.

\begin{figure}[t]
  \centering
  \includegraphics[width=\linewidth]{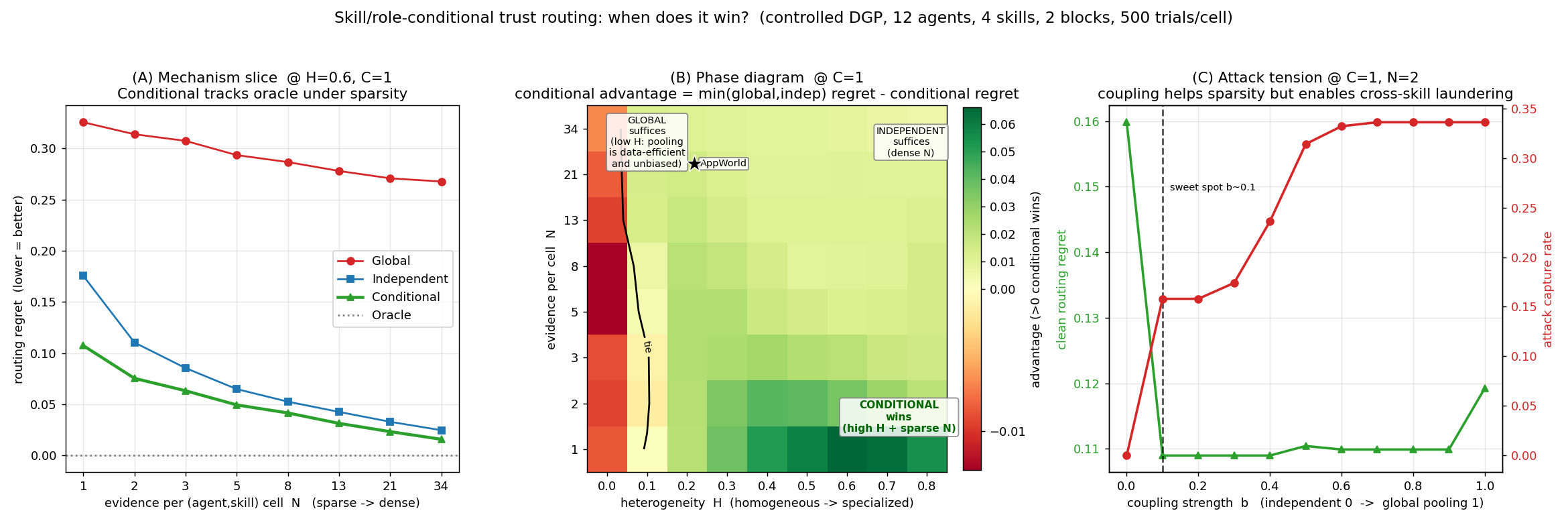}
  \caption{Mechanism (controlled DGP). (A) At fixed heterogeneity, the
  Conditional estimator attains the lowest routing regret when evidence is
  sparse, converging to Independent as $N$ grows. (B) Phase diagram over
  $H\times N$ at $C{=}1$: green = Conditional net-wins; the tie
  line marks where it stops beating the best baseline. Panel C uses
  a perfectly correlated honest farm--target pair ($C{=}1$) with sparse target
  evidence; the AppWorld star in panel B previews the real-data placement expanded
  in Fig.~\ref{fig:overlay}. (C) Attack tension: as coupling $\beta$ rises, clean
  regret falls but laundering capture rises---the dual-use of the borrowing
  channel.}
  \label{fig:phase}
\end{figure}

\begin{figure}[t]
  \centering
  \includegraphics[width=\linewidth]{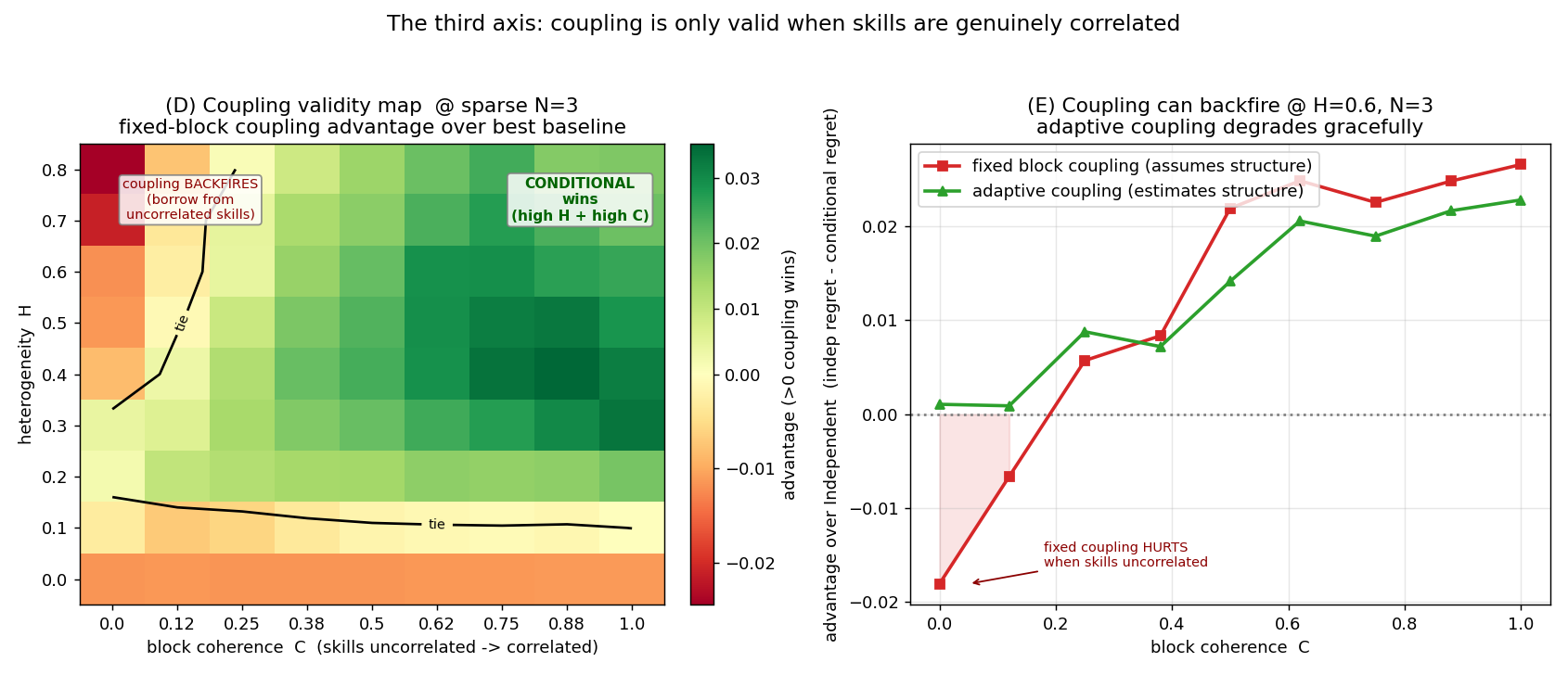}
  \caption{Whether to borrow depends on skill correlation $C$. (D) Coupling
  validity over $C\times H$. (E) Fixed-block coupling \emph{hurts} at low $C$
  (injects irrelevant bias), while adaptive coupling degrades gracefully to
  Independent and stays non-negative across $C$.}
  \label{fig:corr}
\end{figure}

\paragraph{Proposition 1 (When).} \emph{Conditional trust wins exactly under high
$H$ and sparse $N$.} The reason is a bias--variance trade-off read through the
two baselines. Global ($\mat{W}{=}\Ones\Ones^{\top}$) has low variance but is
\emph{biased} whenever agents are specialized---it broadcasts one score across
skills the agent is unequal on---so its error grows with $H$. Independent
($\mat{W}{=}\mat{I}$) is unbiased but \emph{high-variance} when $N$ is small,
because each per-skill estimate rests on a handful of episodes. Conditional
coupling occupies the gap: it borrows correlated cross-skill evidence to cut the
variance Independent suffers, without inheriting the cross-skill bias Global
suffers. The win is therefore largest precisely where both baselines are weak---high
$H$ \emph{and} sparse $N$. Scoring the three $\mat{W}$ choices by
regret~\eqref{eq:regret} (Fig.~\ref{fig:phase}A--B): at $H{=}0.6,N{=}1$ the
Conditional estimator [Eq.~\eqref{eq:pool} with block-$\beta$ $\mat{W}$] attains
regret $0.107$ versus Independent's [$\mat{W}{=}\mat{I}$] $0.176$ ($-40\%$); as
$N$ grows, the borrowing terms in Eq.~\eqref{eq:pool} add little over the
already-abundant diagonal evidence and the two converge. At low $H$, Global
suffices; at high $H$ with dense $N$, Independent suffices. The tie
line in Fig.~\ref{fig:phase}(B) is where Conditional's regret~\eqref{eq:regret}
stops beating the best baseline---the operational boundary CIVT estimates from
logs alone.

\paragraph{Proposition 2 (How much).} \emph{The borrowing strength $\beta$ has a
constrained optimum because the evidence-borrowing channel is also the
reputation-laundering channel.} Here $\beta$ is the off-diagonal scale of
$\mat{W}$ in Eq.~\eqref{eq:pool}: raising it pulls each skill's estimate toward
its block-mates. In Fig.~\ref{fig:phase}(C), clean regret~\eqref{eq:regret} falls
from $0.160$ to $0.109$ for $\beta\!\ge\!0.1$; but the \emph{same} off-diagonal
coupling lets an attacker's farm-skill evidence flow into a target skill through
Eq.~\eqref{eq:pool}, and laundering capture rises from $0$ to $0.34$ with $\beta$
(\S\ref{sec:attack}). The two curves move in opposite directions in the same
parameter, so $\beta$ cannot be set by data efficiency alone: there is a smallest
value, $\beta\!\approx\!0.1$, that secures essentially all of the regret reduction
while sitting at the low end of the laundering-capture ramp. We adopt
$\beta{=}0.1$ as the deployed coupling throughout, and \S\ref{sec:attack} shows
even this conservative setting is enough for an attacker to capture a
\textsc{green} pool. \emph{This dual-use of one knob is the hinge to
\S\ref{sec:attack}.}

\paragraph{Proposition 3 (Whether).} \emph{Whether to borrow depends on $C$, and
the coupling must be estimated, not assumed.} Contrast two ways of setting
$\mat{W}$ in Eq.~\eqref{eq:pool}: fixed block coupling, which \emph{assumes} the
block structure, versus adaptive coupling
$W_{k,k'}{=}\beta\max(R_{k,k'},0)$ built from the estimated correlation
$\mat{R}$~\eqref{eq:corr}. In Fig.~\ref{fig:corr}, fixed coupling \emph{hurts}
when $C$ is low (advantage $-0.018$ at $C{=}0$), because it injects evidence from
uncorrelated skills---borrowing where there is nothing legitimate to borrow. The
adaptive $\mat{W}$ instead stays $\ge 0$ across all $C$, collapsing to $\mat{I}$
when $\mat{R}\!\approx\!\mat{0}$, so it never backfires: borrowing is safe only
along empirically estimated correlation. This makes adaptive coupling the natural
\emph{safe default}---but the safety is conditional on $C$ being low where it
matters. We flag now, and demonstrate in \S\ref{sec:defense}, that the guarantee
inverts under attack: on a genuinely \emph{high}-$C$ pool the adaptive estimator
borrows along exactly the channel the attacker rides, so graceful degradation at
low $C$ does not imply robustness at high $C$.

%% ===========================================================================
\section{Real-Data Placement}
\label{sec:realdata}

The mechanism analysis tells us \emph{where} conditioning wins; it cannot tell us
whether real agent pools fall there. The phase diagram is only useful if real
pools occupy its interesting regions rather than collapsing into a corner. We
therefore take a pool we did not design---a public leaderboard---measure its
$(H,N,C)$, and ask CIVT whether conditioning carries value.

\paragraph{Data.} Among the many benchmarks now used to evaluate LLM
agents~\cite{liu2024agentbench}, we use public AppWorld~\cite{trivedi2024appworld}
because it pairs genuinely different scaffold$\times$model agents with
program-verifiable, per-app outcomes---exactly the heterogeneous, verifiable
evidence our setting needs. We take its leaderboard
bundles: \textbf{14 genuinely heterogeneous agents}---the cross product of
scaffolds \{ReAct~\cite{yao2023react}, PlanExec, FullCodeRefl, IPFunCall\} and base
models \{GPT-4o, GPT-4~Turbo~\cite{openai2023gpt4},
DeepSeekCoder~\cite{guo2024deepseekcoder},
LLaMA3-70B~\cite{grattafiori2024llama3}\}---evaluated on the same $168$
\texttt{test\_normal} tasks ($2352$ episodes). The skill axis is the primary app
of each task (file\_system, phone, simple\_note, splitwise, spotify, todoist,
venmo). No model is executed: we download, parse, and analyze published
trajectories, so the pool is one we did not construct and cannot tune in our
favor---a deliberately adversarial-to-ourselves test of whether real public pools
contain skill-conditional value. The heterogeneity is real by construction: the
scaffold $\times$ base-model cross product mixes planning styles and underlying
models, the two axes most likely to produce skill specialization.

\begin{figure}[t]
  \centering
  \includegraphics[width=\linewidth]{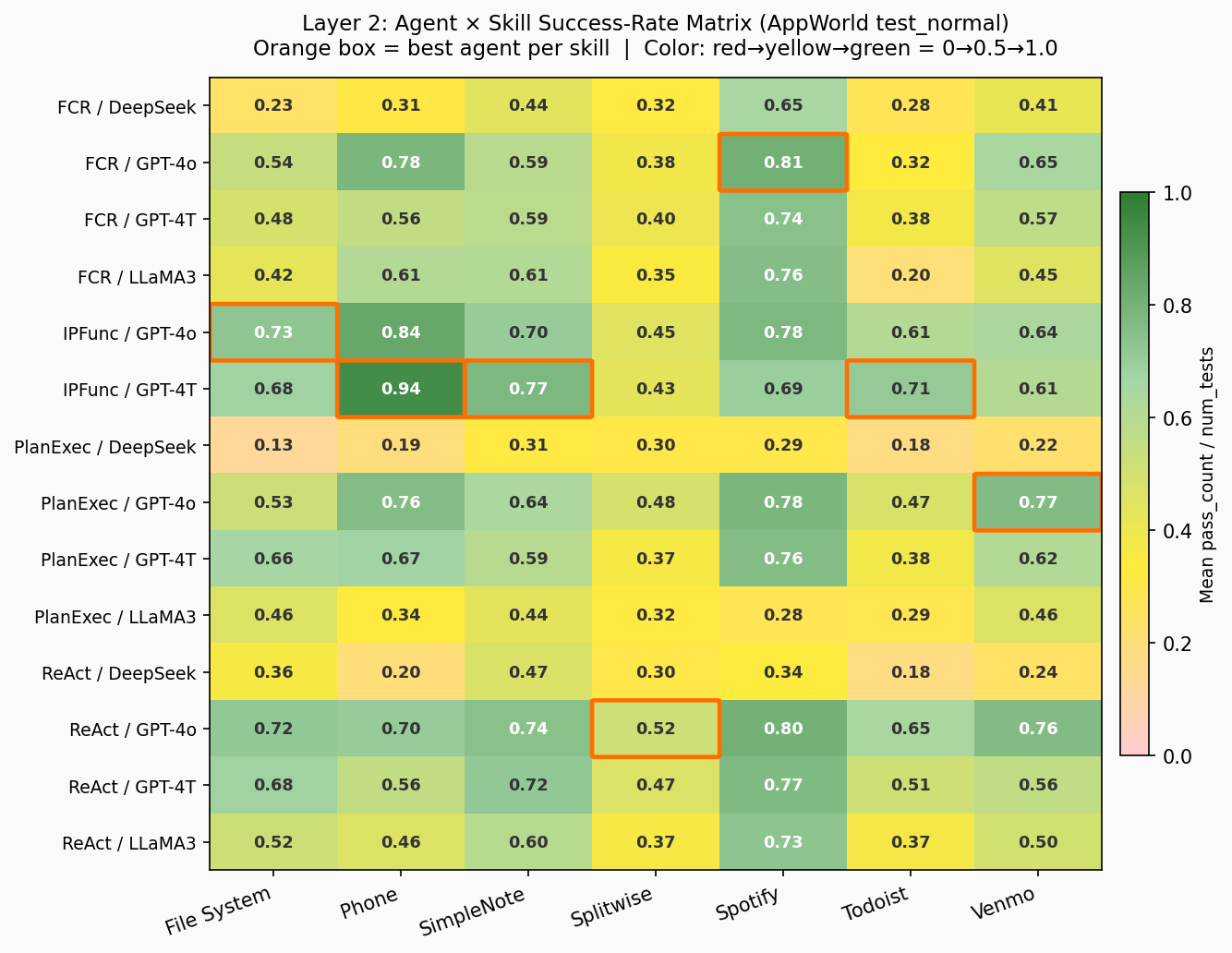}
  \caption{Agent$\times$skill success matrix on AppWorld (14 agents $\times$ 7
  app-skills). Orange boxes mark the best agent per skill: the winner
  \emph{changes across skills}, the basic precondition for skill conditioning to
  carry value.}
  \label{fig:heatmap}
\end{figure}

\paragraph{Result: \textsc{green}.} Over the 14-agent pool
(Table~\ref{tab:routers}), skill conditioning beats the global-best agent by
$\Delta_{\mathrm{sk}}=+0.041$ (score) / $+0.054$ (success), against a total
per-task headroom of $\Delta_{\mathrm{or}}=0.190/0.310$. Two facts make this a
genuine---if modest---win. First, the gain is positive on \emph{both} a
continuous score and a binary success metric, so it is not an artifact of one
scoring choice. Second, the best agent genuinely differs across skills: the best
on \emph{phone} is IPFunCall\slash GPT-4T, on \emph{venmo} PlanExec\slash GPT-4o,
and on \emph{spotify} FullCodeRefl\slash GPT-4o (Fig.~\ref{fig:heatmap}), so the
non-uniqueness condition of CIVT holds and no single global pick is optimal. Both
gaps clear the \textsc{green} thresholds ($\dsk\!\ge\!0.03$, $\dor\!\ge\!0.05$);
the verdict is \textsc{green} on both metrics.

\begin{table}[t]
  \centering
  \caption{Idealized router values on AppWorld (14-agent pool).}
  \label{tab:routers}
  \begin{tabular}{lcc}
    \toprule
    Router & Score & Success \\
    \midrule
    $\rglobal$ & 0.743 & 0.488 \\
    $\rskill$  & 0.784 & 0.542 \\
    $\roracle$ & 0.933 & 0.798 \\
    \midrule
    $\dsk$ (skill vs.\ global)  & $+0.041$ & $+0.054$ \\
    $\dor$ (oracle vs.\ global) & $+0.190$ & $+0.310$ \\
    \bottomrule
  \end{tabular}
\end{table}

\paragraph{Landing point.} The empirical $(H,N,C)=(0.217,\,24,\,0.795)$,
previewed by the AppWorld star in Fig.~\ref{fig:phase}B and expanded with
additional real landings in Fig.~\ref{fig:overlay}, projects into
the shallow green part of the reference $H\times N$ slice; its full
$(H,N,C)$ coordinates explain the modest effect size exactly as
Proposition~1 predicts. $H{=}0.22$ is only moderate: frontier models
on a shared benchmark are more alike than not, limiting the Global baseline's
bias. $N{=}24$ is not sparse: with two dozen episodes per cell, Independent is
already a decent estimator, so there is little variance for borrowing to cut.
$C{=}0.79$ is high: the skills are strongly correlated, so we cannot attribute
the gain to fully orthogonal competence. \emph{Honest reading:} real public pools
do contain skill-conditional value, and the sign and significance survive scrutiny,
but the magnitude is small because AppWorld sits near---not deep inside---the
beneficial regime. We read the small magnitude not as a weak effect but as a
\emph{measured location} on the phase diagram, which converts into a falsifiable
prediction: the gain is small \emph{because} today's public pools are built from a
few frontier models that are mutually similar (moderate $H$, high $C$) and densely
evaluated (large $N$); as open marketplaces fill with more specialized, unevenly
evidenced agents---lower $C$, higher $H$, sparser $N$---a pool's coordinates move
along the diagram toward the deep-green region, and the conditional gain grows with
them. The mechanism is the same; only the substrate's coordinates change. This is
the result (the mechanism is real on real data) and its caveat (the public
landscape is shallow) at once, stated as a curve the ecosystem will climb rather
than a constant. The high $C$ also foreshadows the adversarial result: it is
precisely the correlation that makes borrowing legitimate here that an attacker
will later exploit (\S\ref{sec:attack}).

\subsection{Robustness: the phase diagram holds in both directions}
\label{sec:robustness}
\begin{figure}[t]
  \centering
  \includegraphics[width=\linewidth]{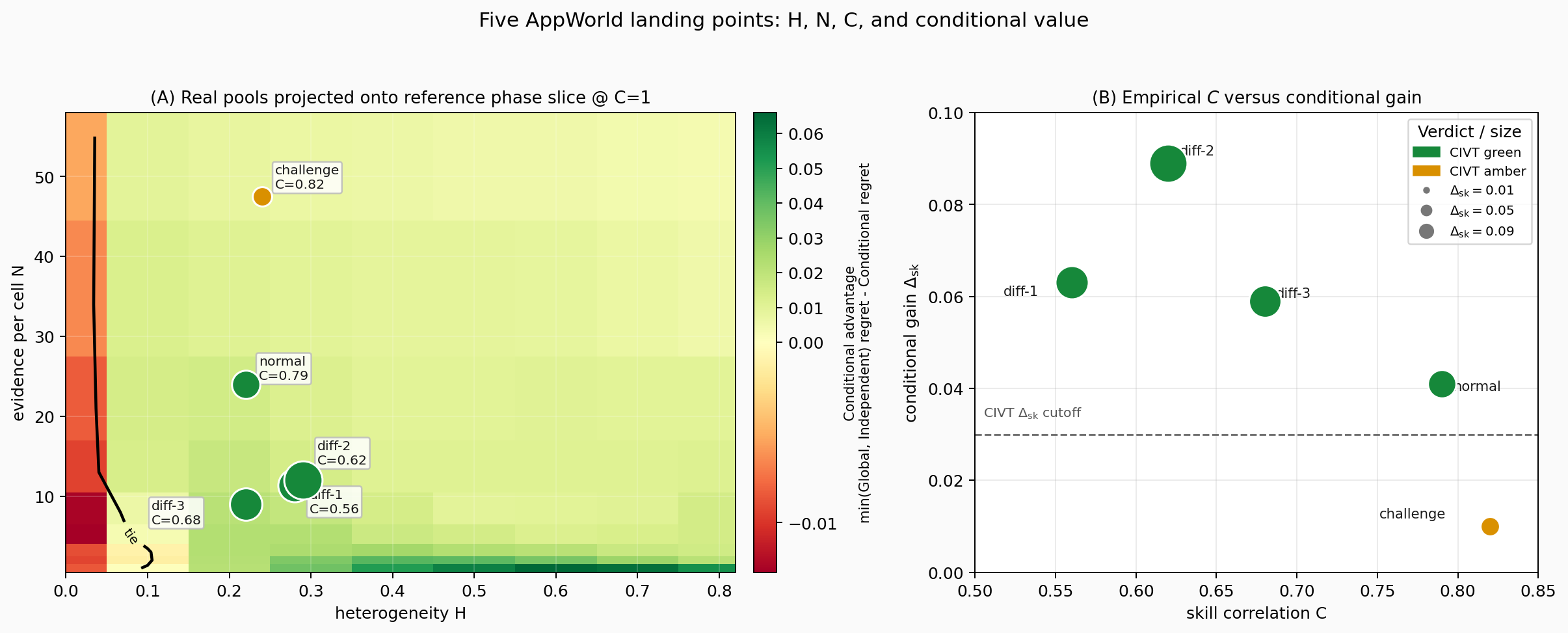}
  \caption{Five real AppWorld landing points. (A) Projection onto a
  reference $H\times N$ phase slice at $C{=}1$; marker color shows the CIVT
  verdict, marker size tracks $\Delta_{\mathrm{sk}}$, and labels report each
  pool's empirical $C$. Since $C$ is not an axis in this panel, (B) plots
  empirical skill correlation $C$ directly against conditional gain
  $\Delta_{\mathrm{sk}}$; the dashed line marks the CIVT skill-value cutoff
  $\Delta_{\mathrm{sk}}{=}0.03$.}
  \label{fig:overlay}
\end{figure}

A single landing point cannot show that the phase diagram \emph{governs} real
pools rather than merely admitting one. We add four more landing points from the
same public source, at zero model cost: three difficulty strata of
\texttt{test\_normal} (slicing the 14-agent pool by AppWorld task difficulty), and
the independent, harder \texttt{test\_challenge} split ($380$ tasks across the same
$14$ agents, recovered by decrypting the public leaderboard bundles; same
app-primary skill axis). Table~\ref{tab:landings} reports the full
coordinates; Fig.~\ref{fig:overlay} visualizes them by projecting $H,N$ onto the
reference phase slice and plotting $C$ directly against the observed conditional
gain. Slicing out difficulty
lowers the skill correlation $C$ from $0.79$ to $0.56$--$0.68$ and the conditional
gain \emph{rises} to $+0.06$--$+0.09$: pooling difficulties had averaged
specialization away, so the headline $+0.04$ \emph{understates} the per-stratum
value. The harder \texttt{test\_challenge} split moves the other way---$C$ rises to
$0.82$ as multi-app tasks and floored-out weak agents inflate cross-skill
correlation---and conditioning value collapses to $+0.01$, an \textsc{amber}
verdict: ample total headroom (oracle$-$global $=0.25$) that the primary-app axis
fails to capture. The governing variable is the correlation $C$ (and the coarseness
of the skill axis), not task difficulty. Crucially, CIVT \emph{discriminates}: it
certifies \texttt{test\_normal} \textsc{green} and flags \texttt{test\_challenge}
\textsc{amber} from logs alone, returning a negative exactly where conditioning
stops paying. The de-risking test is therefore not a rubber stamp---which is what
makes its \textsc{green} on \texttt{test\_normal}, and the attack we now mount on
that pool, meaningful.

\begin{table}[t]
  \centering
  \caption{Five real AppWorld landing points (no model execution).
  The table reports the full $(H,N,C)$ coordinates behind Fig.~\ref{fig:overlay};
  CIVT returns \textsc{green} or \textsc{amber} accordingly.
  $\Delta_{\mathrm{sk}}$ is skill-vs-global on the continuous score.}
  \label{tab:landings}
  \begin{tabular}{lccccc}
    \toprule
    Pool & $H$ & $N$ & $C$ & $\Delta_{\mathrm{sk}}$ & CIVT \\
    \midrule
    \texttt{test\_normal} (\S\ref{sec:realdata}) & 0.22 & 24.0 & 0.79 & $+0.041$ & \textsc{green} \\
    \quad difficulty 1 & 0.28 & 11.4 & 0.56 & $+0.063$ & \textsc{green} \\
    \quad difficulty 2 & 0.29 & 12.0 & 0.62 & $+0.089$ & \textsc{green} \\
    \quad difficulty 3 & 0.22 & 9.0 & 0.68 & $+0.059$ & \textsc{green} \\
    \texttt{test\_challenge} & 0.24 & 47.5 & 0.82 & $+0.010$ & \textsc{amber} \\
    \bottomrule
  \end{tabular}
\end{table}

%% ===========================================================================
\section{Adversarial Analysis}
\label{sec:attack}

Proposition~2 warned, in the controlled DGP, that the borrowing channel is
dual-use; \S\ref{sec:realdata} then found a real pool whose conditional value
rests on high skill correlation. We now collide the two: we run the laundering
attack on exactly that real, CIVT-certified \textsc{green} pool, and ask how
cheap the attack is, which defenses bound it, and at what residual cost.

\paragraph{Threat model.} In an open pool an attacker registers one or more
agents and seeks to be routed skill-$k_t$ tasks despite having no genuine
competence on the \emph{target} skill $k_t$. Its lever is a cheap \emph{farm}
skill $k_f$ on which honest evidence is easy to produce, and the borrowing
channel of Eq.~\eqref{eq:pool}, which lets farm evidence inflate the target
estimate whenever $W_{k_t,k_f}{>}0$. We hold the attacker to an explicit
\textbf{budget} $B$---the number of farm episodes it can fabricate, or the number
of Sybil identities it can register---and the defender routes with the
conditional estimator at the deployed coupling $\beta{=}0.1$ fixed in
\S\ref{sec:mechanism}. We distinguish four attacker profiles by how they spend
the budget: a \emph{launderer} (one agent, perfect farm evidence, zero target
evidence); a \emph{whitewasher} (a fresh identity that sheds a bad history and
re-enters with farm evidence only); a \emph{Sybil} (the same farm budget split
across $k$ identities); and a \emph{sleeper} (farm build-up plus partial,
genuine target evidence). A fifth profile, the \emph{learner}, plants failing
target episodes to defeat the defense of \S\ref{sec:defense} and is analyzed
there. We measure attack benefit by the same full-pool routing
regret~\eqref{eq:regret} used for the honest analysis: the attack succeeds to the
extent it makes the router send target tasks to an incompetent agent.

\paragraph{Attack surface.} The choice of farm--target pair matters for the
strength of the claim. We first examined simple\_note$\rightarrow$venmo, but its
honest gap is sub-\textsc{green} ($\dsk{=}0.02$): CIVT would not recommend
conditioning on that pair in the first place, so capturing it invites the
reply ``you attacked a pool the system would avoid.'' To close that gap we attack
a pair CIVT itself \emph{endorses}: farm\,=\,simple\_note, target\,=\,phone. This
pair is highly correlated ($R{=}0.861$)---so the honest borrowing the attacker
abuses is exactly the borrowing Proposition~3 sanctions---and \textsc{green} under
the honest gated verdict, $\dsk{=}\dor{=}0.1922$, well above the $0.03/0.05$
thresholds. The attack is therefore not an edge case on a pair the system would
avoid; it captures a skill-conditioned pool with real, CIVT-certified routing
value. This is the strongest form of the threat: the defense cannot escape it by
simply refusing to condition, because refusing to condition here forfeits a
genuine $+0.19$ gain.

\begin{figure}[t]
  \centering
  \includegraphics[width=\linewidth]{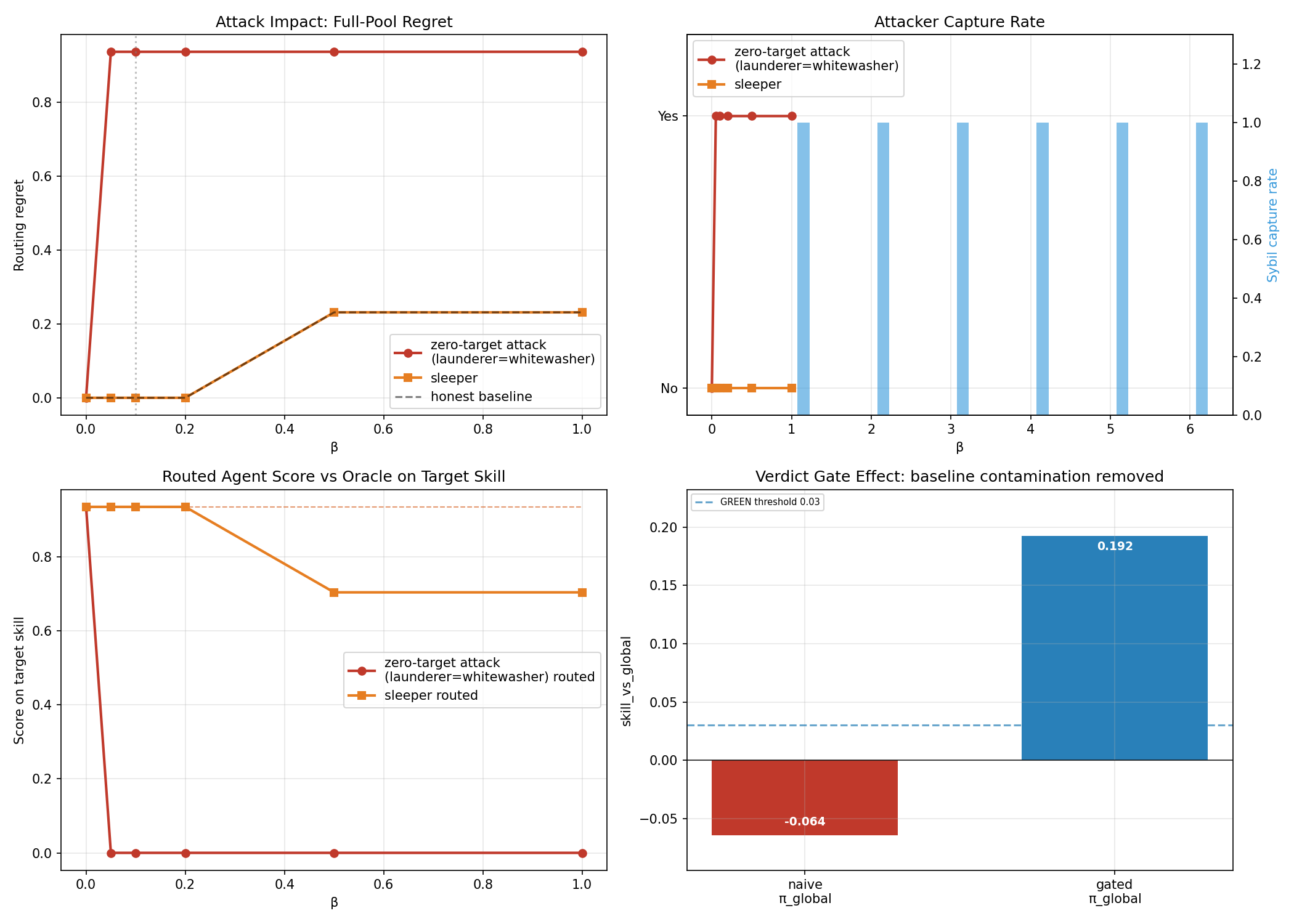}
  \caption{Cross-skill laundering on a \textsc{green} AppWorld pair
  (simple\_note$\rightarrow$phone). As $\beta$ crosses $0.05$, the zero-target
  attack captures the target skill and regret jumps to $0.9357$. The gated
  verdict remains strongly \textsc{green} ($+0.1922$), while the naive global
  baseline is contaminated ($-0.0643$).}
  \label{fig:attack}
\end{figure}

Sweeping $\beta$ (Fig.~\ref{fig:attack}, Table~\ref{tab:attack}), a
\textbf{zero-target attacker}---either a launderer with perfect farm evidence or
a whitewasher with a fresh identity---is captured at $\beta\!\ge\!0.05$ and drives
full-pool routing regret~\eqref{eq:regret} from $0$ to $0.9357$: above the
threshold coupling the inflated target estimate exceeds every honest incumbent's,
so the router hands phone tasks to an agent with no phone competence. The same
attack contaminates the trust signal itself. On the clean pool the gated CIVT
verdict is $+0.1922$ (strongly \textsc{green}); but the launderer's perfect farm
score also lifts its \emph{global} mean, so if it is admitted into the global
baseline---as a naive verdict does---the apparent value of conditioning collapses
to $-0.0643$. The same pool thus reads ``conditioning is harmful'' to a defender
who does not gate, and ``conditioning is worth $+0.19$'' to one who does: the gap
between $-0.06$ and $+0.19$ is entirely an artifact of whether zero-target
evidence is allowed to vote, which is exactly what the defense of
\S\ref{sec:defense} controls. The remaining profiles confirm the pattern: a
\textbf{Sybil} splitting $24$ farm episodes across $k$ identities is captured for
all $k\!\ge\!1$ (splitting the budget does not weaken it, since each identity
still carries zero target evidence), while a \textbf{sleeper} that mixes in
partial genuine target evidence is \emph{not} captured at $\beta{=}0.1$---its real
target failures dilute the laundering signal below the capture threshold. This
realizes Proposition~2's warning on a real \textsc{green} pool, at the
conservative coupling $\beta{=}0.1$ we adopted there.

\begin{table}[t]
  \centering
  \caption{Capture at the $\beta{=}0.1$ sweet spot on the
  simple\_note$\rightarrow$phone \textsc{green} pair.}
  \label{tab:attack}
  \begin{tabular}{lccc}
    \toprule
    Attack & Captured & Min.\ $\beta$ & Farm evidence \\
    \midrule
    Launderer    & yes & 0.05 & 24 ep (1 full agent) \\
    Whitewasher  & yes & 0.05 & 8 ep (fresh identity) \\
    Sybil ($k{=}1$--6) & yes & 0.05 & 24 ep total \\
    Sleeper      & \textbf{no} & --- & 24 farm + target evidence \\
    \bottomrule
  \end{tabular}
\end{table}

\subsection{The attack is one-episode cheap}
A natural first instinct is to defend by rate-limiting: cap how much farm evidence
$B$ an agent may accumulate, on the theory that laundering needs a large farm
balance to outweigh honest competence. We measure attack benefit as full-pool
routing regret~\eqref{eq:regret} against $B$ and find this instinct is
\emph{mathematically void} for pure laundering.

\begin{figure}[t]
  \centering
  \includegraphics[width=\linewidth]{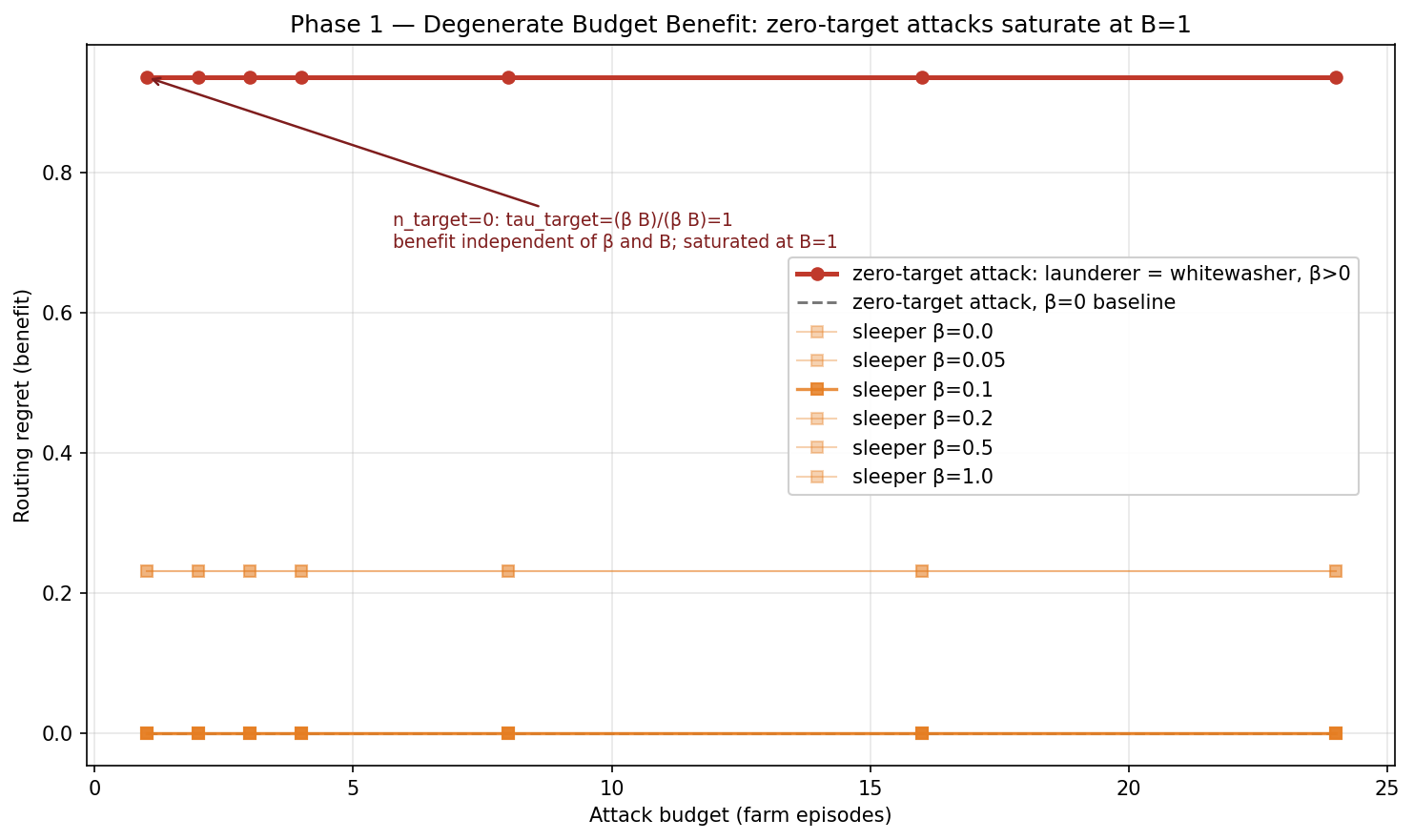}
  \caption{Attack benefit vs.\ budget $B$ on the \textsc{green} pair. For
  zero-target launderer/whitewasher attacks, $B$ and $\beta$ cancel in
  Eq.~\eqref{eq:pool}: any $B\!\ge\!1$ and $\beta\!>\!0$ yields the same capped
  regret ($0.9357$). The sleeper does not capture at $\beta{=}0.1$.}
  \label{fig:budget}
\end{figure}

\paragraph{Budget degeneracy.} Consider a zero-target launderer on target skill
$k_t$ with farm skill $k_f$: it holds no genuine target evidence
($n_{i,k_t}{=}0$) and farm evidence equal to its budget ($n_{i,k_f}{=}B$). Every
honest target term then drops out of Eq.~\eqref{eq:pool}, and the target estimate
collapses:
\[
\hat\tau_{i,k_t}
  = \frac{\beta\,B\,o_{i,k_f} + \overbrace{n_{i,k_t}\,o_{i,k_t}}^{=\,0}}
         {\beta\,B + \underbrace{n_{i,k_t}}_{=\,0}}
  = \frac{\beta\,B\,o_{i,k_f}}{\beta\,B}
  = o_{i,k_f},
\]
identically for \emph{every} $B\!\ge\!1$ and \emph{every} $\beta\!>\!0$: the
budget $B$ and the coupling $\beta$ cancel algebraically. The estimate is pinned
at the farm score the instant a single farm episode exists, so one fabricated
episode already drives regret to its capped value $0.9357$ and a full agent's
$24$ episodes buy nothing more (Fig.~\ref{fig:budget}). The flat budget curve is
not a plotting artifact---it is this cancellation. Two consequences follow.
\emph{(i)} The attack is not merely possible but \emph{one-episode cheap}: there
is no budget threshold to raise, because the threshold is one. \emph{(ii)} The
defense cannot be a rate limit on $B$ or a clip on $\beta$, both of which the
cancellation neutralizes; it must instead be \emph{structural}, keying on the
qualitative fact that genuine target evidence is absent. That is precisely the
gate of \S\ref{sec:defense}.

\subsection{Only an evidence gate bounds it}
\label{sec:defense}
\begin{figure}[t]
  \centering
  \includegraphics[width=\linewidth]{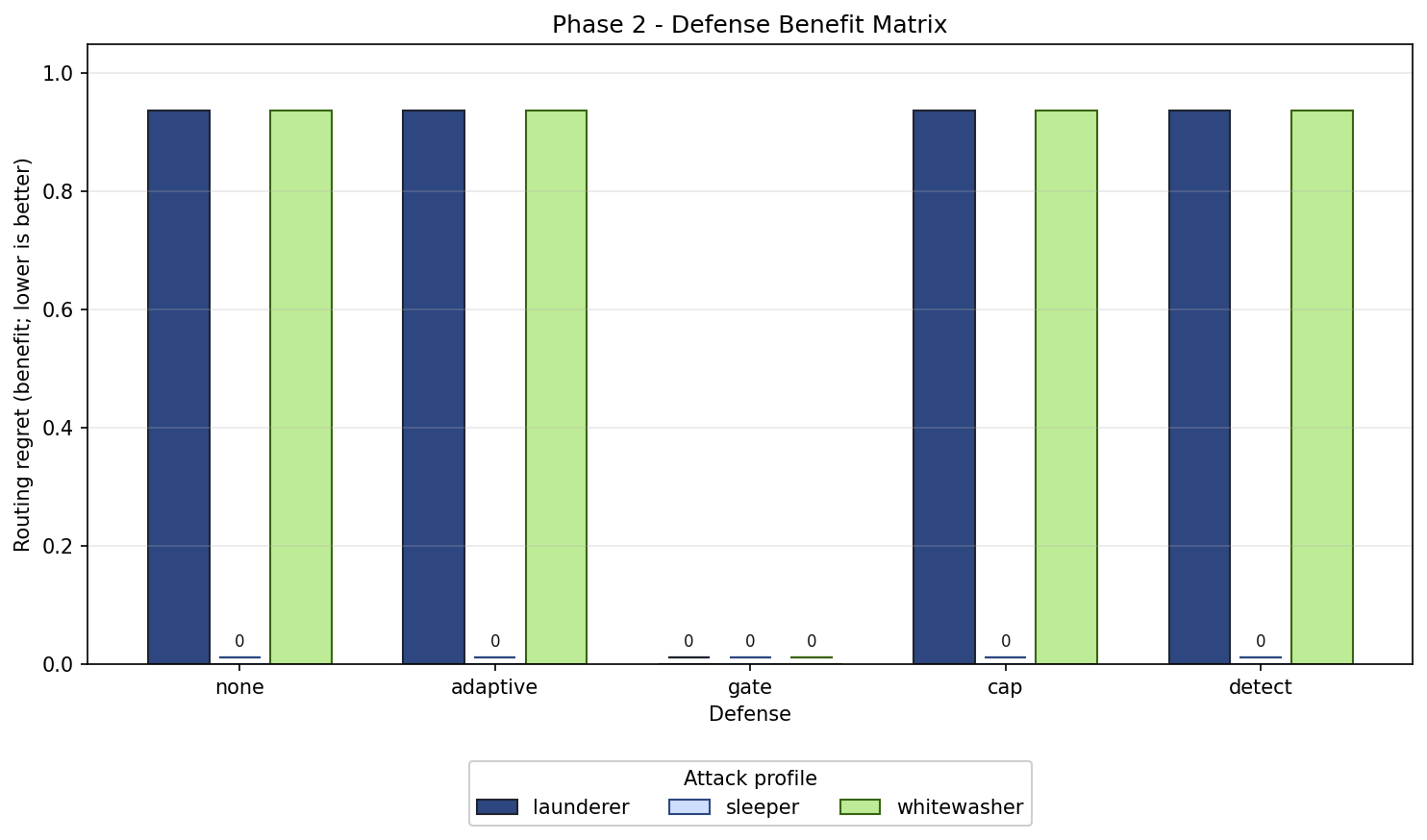}
  \caption{Defense benefit matrix at $\beta{=}0.1$ (lower is better). Of four
  defenses, only the zero-evidence gate reduces launderer and whitewasher regret
  to $0$; adaptive coupling, capping, and detection leave the $0.9357$
  benefit intact. Zero-height bars are marked explicitly at the baseline. The
  sleeper is harmless throughout.}
  \label{fig:defense}
\end{figure}

Budget degeneracy already rules out rate-limiting, but three other defensive
instincts remain: make the \emph{estimator} robust (adaptive coupling), \emph{cap}
the borrowed contribution, or \emph{detect} and flag the attacker. We evaluate all
four against each attack profile at the deployed $\beta{=}0.1$
(Fig.~\ref{fig:defense}), and the picture is sharply asymmetric---only the one
structural defense works:
\begin{itemize}
\item \textbf{Zero-evidence gate}---suppress cross-skill borrowing for any agent
with $n_{i,k}{<}1$ direct episodes on the target skill---is the \emph{only}
effective defense, cutting launderer and whitewasher benefit from $0.9357$ to
$0.0000$ while leaving the honest \textsc{green} verdict intact.
\item \textbf{Adaptive coupling}---the hero of Proposition~3---\emph{fails} here.
Because the farm and target skills are genuinely correlated in the honest pool
($R{=}0.861$), the adaptive estimator~\eqref{eq:corr} borrows along exactly the
channel the attacker rides, and cannot separate laundering from honest correlated
competence. Graceful degradation at \emph{low} $C$ (\S\ref{sec:mechanism}) does
not imply robustness at \emph{high} $C$.
\item \textbf{Capped borrowing} (clip the off-diagonal contribution) is
ineffective at this $\beta$: the coupling structure of Eq.~\eqref{eq:pool} still
admits full inflation.
\item \textbf{Detection signature} (flag agents with perfect farm and zero target
evidence) attains precision and recall $1.0$ at capture, but flagging is not
mitigation:
benefit is unchanged unless the flag triggers an action such as the gate.
\end{itemize}

\paragraph{Bounded claim, and where it breaks.} We therefore state a deliberately
bounded result: \emph{at $\beta{=}0.1$, the zero-evidence gate reduces zero-target
laundering benefit from $0.9357$ to $0$ on a \textsc{green} pair, while leaving the
honest $+0.1922$ verdict intact.} The gate works because it keys on the one thing
budget degeneracy could not fake: the \emph{qualitative} presence of genuine
target evidence. A pure launderer has $n_{i,\mathrm{target}}{=}0$ by definition, so
suppressing cross-skill borrowing exactly for $n_{i,k}{<}1$ removes its inflation
without touching any agent that has actually attempted the target skill. But this
is a bounded result, not Sybil-resistance, and we are explicit about where it
breaks. The predicate $n_{i,k}{<}1$ is a cliff at zero, so an attacker who plants
even a single failing target episode---posing as a ``learner'' acquiring the skill
rather than a pure launderer---steps over the cliff and re-enters the borrowing
channel. The question is what that costs.

\begin{figure}[t]
  \centering
  \includegraphics[width=\linewidth]{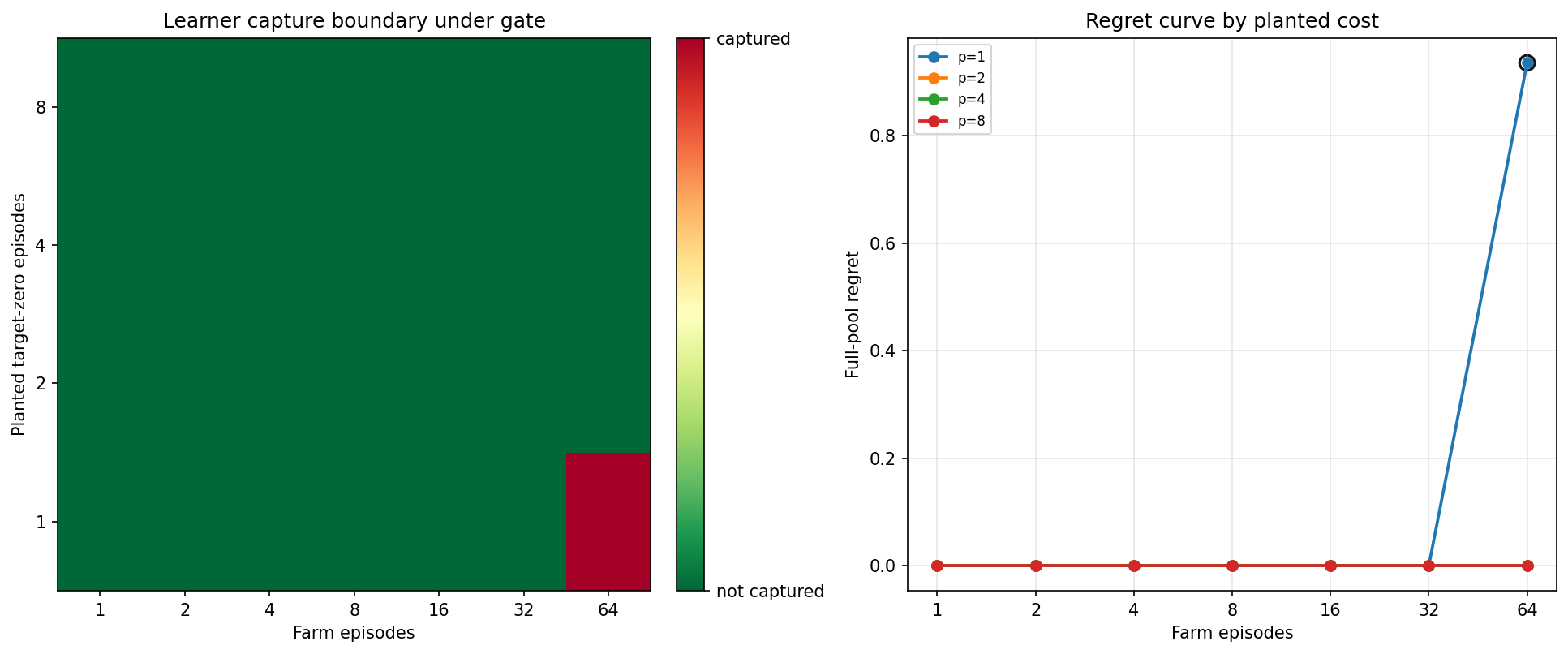}
  \caption{Residual learner cost under the zero-evidence gate. Once the attacker
  plants target-skill evidence, the budget axis becomes non-degenerate: on the
  \textsc{green} pair, one planted failing target episode requires $64$ farm
  episodes to recover capture, while two or more planted episodes are not
  captured up to $64$ farm episodes.}
  \label{fig:learner}
\end{figure}

Figure~\ref{fig:learner} quantifies this residual channel, and the cost is high.
The budget axis, degenerate for pure laundering, comes back to life once a planted
episode appears in the denominator of Eq.~\eqref{eq:pool}: the attacker must now
out-mass its own planted failures with farm evidence. With one planted
target-zero episode, regaining capture takes $64$ farm episodes; with two, four,
or eight planted episodes, capture does not occur at all up to $64$ farm episodes.
The gate is therefore not a cure, but it converts a free, one-episode attack into
an explicit and steep target-evidence-versus-farm-budget tradeoff. Crucially, the
bypass is \emph{costlier on greener pools}: a high-$\dsk$ pool has strong honest
incumbents, so each planted failure drags the attacker's pooled estimate below
their bar faster than farm evidence can lift it back---the gate is most effective
exactly where conditioning is most valuable, and weakest on the marginal pools
where there was little to protect. This is the shape of the bounded-defense
claim we make: rather than assert Sybil-resistance, we map which attacks each
defense blocks, at what cost, and where the classical Sybil-proofness
impossibility~\cite{cheng2005sybilproof,friedman2001whitewashing} remains open---a
trade-off quantified, not a guarantee overclaimed.

%% ===========================================================================
\section{Limitations and Future Work}
\label{sec:limitations}
We state the boundaries of the claims plainly.
\begin{itemize}
\item \textbf{Verifiable-outcome scope.} We study skills whose outcome a program
can check---tool use, code, payments, data tasks (the makeup of AppWorld). For
\emph{quality} services such as summarization, writing, or translation there is no
environment judge, and the only available signal is a consumer's \emph{subjective}
rating; such skills fall outside our setting. When unverifiable
peer ratings enter the signal, transitive cross-agent propagation (and the
Sybil/collusion surface it must then defend) becomes a live problem again; we
leave that regime to future work. We view restricting to the verifiable slice as
deliberate, not a simplification: when a checkable signal exists, routing on it
strictly dominates routing on a gameable rating, and the slice already covers a
large class of real agent tasks.
\item \textbf{Skill definition.} The skill axis is the single primary app of each
task; multi-app tasks are projected onto their primary app, and task
\emph{difficulty} is not modeled as a separate skill. A finer skill taxonomy
could raise both $H$ and the conditional gain.
\item \textbf{Shallow real landing.} AppWorld lands inside the beneficial regime
but shallowly (high $C{=}0.79$, modest gain ${+}0.04$). We validate the mechanism
on real data but cannot, on this benchmark, exhibit a \emph{large} real-world
effect; the difficulty strata of \S\ref{sec:robustness} already show the gain
rising as $C$ falls, but reaching the deep-green regime would require pools with
higher heterogeneity and sparser, less-correlated evidence than current public
benchmarks supply.
\item \textbf{Bounded adversarial claim.} The gate's guarantee is bounded, not
absolute: it defeats zero-target laundering but is bypassable by a learner at a
quantified cost (\S\ref{sec:defense}), and we explicitly do not claim
Sybil-resistance.
\item \textbf{Conditioning axis.} We condition on skill $k$ only; conditioning on
role $r$ and query $q$, and the interaction among them, is future work.
\end{itemize}

%% ===========================================================================
\section{Conclusion}
\label{sec:conclusion}
As open systems fill with heterogeneous LLM agents, a single global trust score
becomes the wrong object, and the natural fix---conditioning trust on the task's
skill---is neither universally better nor a free lunch. Our
characterization makes three things precise. \emph{When}: conditioning pays off
only under high heterogeneity and sparse, correlated evidence, a regime CIVT
detects from logs at zero cost. \emph{How much}: the borrowing strength $\beta$
has a constrained optimum because the same channel that buys data efficiency
launders reputation. \emph{Whether it is safe}: cross-skill laundering captures
even a CIVT-certified \textsc{green} pool at one-episode cost, and only a
structural zero-evidence gate bounds it---bounds, not eliminates, with a residual
learner cost we quantify. We give the map, place real public data on it, and chart
its adversarial limits, trading the comfortable claim that conditional trust is
simply better for the more durable one that says exactly when, how much, and at
what risk.

\bibliographystyle{abbrvnat}
\bibliography{refs}

\end{document}